\pdfoutput=1
\documentclass[11pt]{article}

\usepackage[preprint]{acl}
\usepackage{times}
\usepackage{latexsym}

\usepackage[T1]{fontenc}
\usepackage[utf8]{inputenc}

\usepackage{microtype}

\usepackage{inconsolata}

\usepackage{amsmath, amssymb, bbm, mathrsfs}
\usepackage{amsfonts}
\usepackage{graphicx}
\usepackage{CJKutf8}
\usepackage{booktabs}
\usepackage{multirow}
\usepackage{longtable}
\usepackage{url}
\usepackage{colortbl}

\title{DDO: Dual-Decision Optimization for LLM-Based Medical Consultation via Multi-Agent Collaboration}

\author{
 \textbf{Zhihao Jia\textsuperscript{1}},
 \textbf{Mingyi Jia\textsuperscript{1}},
 \textbf{Junwen Duan\textsuperscript{1}}\thanks{Corresponding Author. Email: \href{mailto:jwduan@csu.edu.cn}{jwduan@csu.edu.cn}.},
 \textbf{Jianxin Wang\textsuperscript{1}},
\\
 \textsuperscript{1}Hunan Provincial Key Lab on Bioinformatics, \\ School of Computer Science and
Engineering, Central South University \\
\\
\texttt{\{zhihaojia, jiamingyi, jwduan\}@csu.edu.cn, jxwang@mail.csu.edu.cn}
}

\begin{document}
\begin{CJK}{UTF8}{gbsn}
\maketitle

\begin{abstract}
Large Language Models (LLMs) demonstrate strong generalization and reasoning abilities, making them well-suited for complex decision-making tasks such as medical consultation (MC). However, existing LLM-based methods often fail to capture the dual nature of MC, which entails two distinct sub-tasks: symptom inquiry, a sequential decision-making process, and disease diagnosis, a classification problem. This mismatch often results in ineffective symptom inquiry and unreliable disease diagnosis. To address this, we propose \textbf{DDO}, a novel LLM-based framework that performs \textbf{D}ual-\textbf{D}ecision \textbf{O}ptimization by decoupling the two sub-tasks and optimizing them with distinct objectives through a collaborative multi-agent workflow. Experiments on three real-world MC datasets show that DDO consistently outperforms existing LLM-based approaches and achieves competitive performance with state-of-the-art generation-based methods, demonstrating its effectiveness in the MC task. The code is available at https://github.com/zh-jia/DDO.

\end{abstract}

\begin{figure}[t]
    \centering
    \includegraphics[width=0.8\columnwidth]{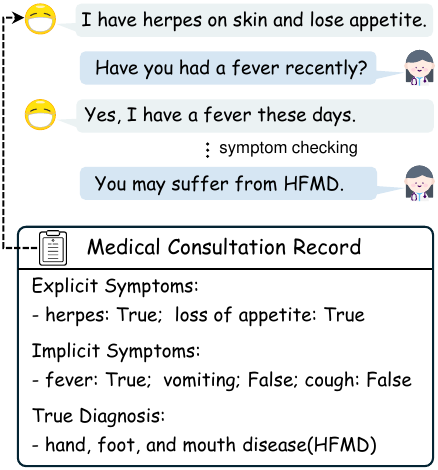}
    \caption{An example of a Medical Consultation (MC) task, where an AI doctor iteratively inquires about additional symptoms based on the patient's initial self-reported symptoms and ultimately provides a diagnosis.}
    \label{fig:example}
\end{figure}

\section{Introduction}
\textbf{Medical Consultation (MC)}, aiming to automate symptom collection and support clinical diagnosis, has become a promising application in AI-driven healthcare and attracted growing attention~\citep{zhao2024haiformer, hu2024uncertainty, chopra2025feedback}. As shown in Figure~\ref{fig:example}, MC involves multi-turn interactions between an AI doctor and a patient, encompassing two core decision-making processes: symptom inquiry—a sequential decision task over a large action space—and disease diagnosis—a classification task over a limited set of candidate diseases~\citep{chen2023dxformer, chen2024cod}. The effectiveness of MC hinges on the AI doctor's ability to perform both efficient information seeking and accurate disease differentiation.

Compared to the models used in traditional reinforcement learning (RL)-based methods~\citep{zhong2022hierarchical, yan2023efficient} and generation-based approaches~\citep{hou2023mtdiag, zhao2024haiformer}, large language models (LLMs) provide stronger generalization and more transparent reasoning~\citep{qin2024enhancing, singh2024deciphering}, potentially reducing training costs and improving interpretability for the MC task. However, due to hallucinations and limited domain adaptation, directly applying LLMs to MC often results in inefficient inquiry and unreliable diagnosis. Recent approaches~\citep{hu2024uncertainty, chopra2025feedback} improve information gathering by incorporating LLM-based planning, yet overlook diagnostic optimization. To jointly optimize symptom inquiry and disease diagnosis,~\citet{chen2024cod} introduced the Chain-of-Diagnosis (CoD) framework, enabling LLMs to learn both components from CoD training data. Nevertheless, the fundamentally different nature of these two decision-making sub-tasks presents significant challenges for unified optimization.

To overcome these limitations, we propose \textbf{DDO}, an LLM-based MC approach that leverages multi-agent collaboration to decouple and optimize the two core decision-making components with distinct objectives. For symptom inquiry, DDO integrates a lightweight RL-based policy agent that generates reliable candidate actions, thereby reducing the decision-making burden on LLMs. For disease diagnosis, DDO derives fine-grained diagnostic confidence from LLM logits and enhances disease discrimination through a plug-and-play adapter trained via in-batch contrastive learning. Experiments on three real-world MC datasets demonstrate that DDO consistently outperforms other LLM-based methods and achieves performance on par with state-of-the-art (SOTA) generation-based approaches, while requiring substantially less training overhead. Our contributions are as follows:
\begin{itemize}
    \setlength{\itemsep}{0.5pt}
    \item We introduce DDO, a novel multi-agent framework for the MC task, where four collaborative agents enable an effective and transparent diagnostic MC workflow. 
    \item DDO decouples the two core decision-making processes—symptom inquiry and disease diagnosis—and optimizes them with distinct objectives, leading to more informative questioning and improved diagnostic accuracy.
    \item By tuning only a small number of model parameters, DDO surpasses other LLM-based methods and achieves performance comparable to SOTA generation-based approaches.
\end{itemize}

\section{Related Work}
\subsection{Medical Consultation Task}
Medical Consultation (MC), a key application of AI in medicine~\citep{valizadeh2022ai}, was initially formulated as a Markov Decision Process (MDP) and optimized using reinforcement learning (RL)~\citep{tang2016inquire, wei2018task, kao2018context}. However, due to the high variability of RL agents~\citep{xia2020generative}, researchers have incorporated disease-symptom prior knowledge to enhance the decision-making~\citep{xu2019end, liu2022my, yan2023efficient} by the agents. HRL~\citep{zhong2022hierarchical} introduced a hierarchical RL framework to refine the action space. Additionally, generative approaches such as Diaformer~\citep{chen2022diaformer}, CoAD~\citep{wang2023coad}, MTDiag~\citep{hou2023mtdiag} and HAIformer~\citep{zhao2024haiformer} leveraged attention mechanisms to enhance optimization efficiency, achieving SOTA performance in the MC task. AIME~\citep{tu2024towards} demonstrated the potential of LLMs in medical history-taking by learning from realistic medical dialogues. MediQ~\citep{li2024mediq} introduced an abstention module to assess whether the collected diagnostic evidence is sufficient. UoT~\citep{hu2024uncertainty} and MISQ-HF~\citep{chopra2025feedback} aimed to reduce decision uncertainty through LLM-driven planning. CoD~\citep{chen2024cod} improved the interpretability of doctor agent's decisions by generating transparent chained thought processes. 

\subsection{LLMs in Medical Decision-Making}
LLMs have demonstrated strong potential across various medical applications~\citep{zhou2024large}. They are capable of answering medical exam questions~\citep{kim2024mdagents, shi2024medadapter}, collecting patient history~\citep{johri2024craft}, offering diagnostic suggestions~\citep{jia2025medikal, rose2025meddxagent}, and recommending treatment plans~\citep{li2024agent}. Leveraging prompt engineering~\citep{zheng2024exploring, liu2024medcot} and domain adaptation techniques~\citep{tian2024chimed, wang2025citrus}, their reasoning capabilities have significantly improved, leading to more reliable medical decision-making. Moreover, to tackle more complex tasks in the medical domain, recent studies~\citep{kim2024mdagents, bani2024magda} have explored the use of multiple LLM agents, offering promising directions for enabling collaborative decision-making in challenging clinical scenarios.

\begin{figure*}[t]
  \includegraphics[width=\linewidth]{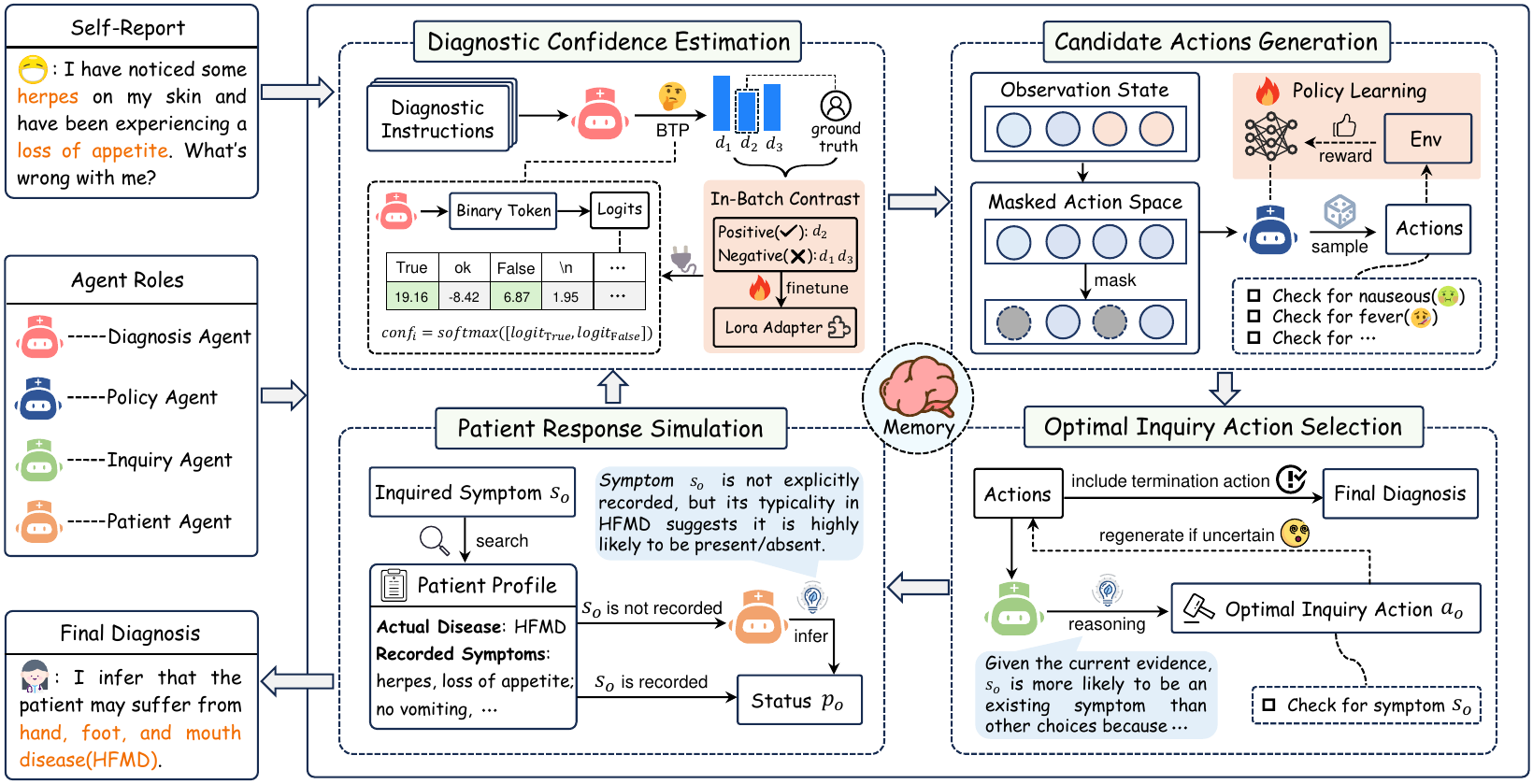}
  \centering
  \caption {Overview of the proposed DDO framework, comprising four collaborative agents operating over a shared memory to execute the consultation workflow: the \textit{Diagnosis Agent} estimates disease confidences from LLM logits; the \textit{Policy Agent} generates candidate actions via masked sampling; the \textit{Inquiry Agent} selects the optimal symptom to query or terminates the consultation; and the \textit{Patient Agent} responds based on the patient profile.}，
  \label{fig:framework}
\end{figure*}

\section{Problem Definition} 
A real-world Medical Consultation Record (MCR) is denoted as $\mathcal{P} = \{ \mathcal{S}^\text{ex}, \mathcal{S}^\text{im}, d_l \}$, where $\mathcal{S}^\text{ex} = \{(s_i^\text{ex}, p_i^\text{ex})\}_{i=1}^{l_1}$ represents \textit{explicit symptoms} initially reported by the patient, and \textit{implicit symptoms} $\mathcal{S}^\text{im} = \{(s_j^\text{im}, p_j^\text{im})\}_{j=1}^{l_2}$ are elicited through follow-up inquiries by the doctor. The label $d_l$ denotes the ground-truth disease of the patient.

The MC task simulates a multi-turn interaction process between an AI doctor and a simulated patient, where the AI doctor actively collects diagnostic information to facilitate differential diagnosis. Starting from the initial symptoms $\mathcal{S}^\text{ex}$, the AI doctor selectively inquires about additional symptoms $\mathcal{S}^\text{ad}$ to accumulate diagnostic evidence. The interaction terminates when sufficient information is collected or a predefined maximum number of turns $L$ is reached. The final diagnosis $d_p$ is selected from the candidate set $D = \{d_1, d_2, \dots, d_n\}$ based on the highest diagnostic confidence:
\begin{equation}
d_p = \arg\max_{d_i \in D} \text{conf}_i\left(\mathcal{S}^\text{ex} \cup \mathcal{S}^\text{ad}, d_i\right)
\label{eq:prediction}
\end{equation}

\section{Multi-Agent Collaborative Medical Consultation in DDO}
  
To enhance the decision-making capability of LLMs in the MC task, the DDO framework integrates three LLM-based agents—\textit{\textit{Diagnosis Agent}}, \textit{Inquiry Agent}, and \textit{Patient Agent}—alongside an RL-based \textit{Policy Agent} implemented with an actor-critic architecture. The \textit{Diagnosis Agent} employs a learnable adapter to enhance the LLM’s ability to perform differential diagnosis, while the \textit{Policy Agent} and the \textit{Inquiry Agent} collaborate to strategically acquire informative symptoms. All agents operate over a shared \textbf{\textit{memory}}, consisting of (1) a static component that encodes prior disease–symptom knowledge exclusively derived from the training portion of the dataset (thus avoiding any leakage of test labels), and (2) a dynamic component that is continuously updated with observed symptoms and diagnostic confidence throughout the consultation process.

As shown in Figure~\ref{fig:framework}, each consultation round proceeds through four steps: 1) \textbf{Diagnostic Confidence Estimation}—The \textit{Diagnosis Agent} estimates confidence scores for each candidate disease based on the current diagnostic evidence. 2) \textbf{Candidate Actions Generation}—The \textit{Policy Agent} samples multiple interaction actions based on the current state, providing a set of reliable choices for the next inquiry. 3) \textbf{Optimal Inquiry Action Selection}—The \textit{Inquiry Agent} selects the most informative symptom checking action from the candidate actions. 4) \textbf{Patient Response Simulation}—The \textit{Patient Agent} simulates the patient’s responses, indicating whether the inquired symptom is present.

\subsection{Diagnostic Confidence Estimation}

\subsubsection{Binary Token Probability} 
The \textit{Diagnosis Agent} estimates the diagnostic confidence score $ conf_i \in (0, 1) $ for each candidate disease $d_i \in D$, reflecting the likelihood of $d_i$ being the correct diagnosis given the current evidence. Rather than relying on decoding to generate numeric scores~\citep{li2024mediq, chen2024cod, qin2024enhancing}, the \textit{Diagnosis Agent} adopts \textbf{Binary Token Probability (BTP)}—a logit-based method inspired by multiple-choice QA~\citep{detommaso2024multicalibration, kumar2024confidence}—to provide a more efficient and interpretable confidence estimation for LLMs.

Specifically, given a structured prompt that integrates the current evidence with disease-specific knowledge of $d_i$ retrieved from the shared memory, the LLM-based \textit{Diagnosis Agent} is instructed to output a single binary token (\texttt{True}/\texttt{False}) indicating whether $d_i$ is a plausible diagnosis. We extract the logits distribution at the position of this binary token and obtain the logits corresponding to \texttt{True} and \texttt{False}, denoted as $\text{logit}_\text{T}$ and $\text{logit}_\text{F}$, respectively. The diagnostic confidence is finally computed via a temperature-scaled softmax over the binary logits, where the temperature $\tau$ controls the sharpness of the logits distribution:
\begin{equation}
conf_i = \frac{\exp(\text{logit}_\text{T} / \tau)}{\exp(\text{logit}_\text{T} / \tau) + \exp(\text{logit}_\text{F} / \tau)}
\label{eq:btp}
\end{equation}

\subsubsection{Calibrating the Diagnostic Confidence}
Diagnostic confidence scores from base LLMs often lack discriminative power when candidate diseases share overlapping symptoms—e.g., both \textit{upper respiratory tract infections} and \textit{pneumonia} commonly present with \textit{fever} and \textit{cough}. Existing calibration methods typically require fine-grained supervision~\citep{detommaso2024multicalibration, chen2024cod}, such as expert-annotated confidence scores, which are often infeasible to implement in clinical practice. Instead, we treat diagnosis as a multi-class classification task~\citep{ma2024memorize} and leverage ground-truth disease labels $d_l$ from Medical Consultation Records (MCRs) as weak supervision to calibrate the diagnostic confidence.

To construct the calibration training data, we generate partial consultation trajectories by truncating each full MCR at different interaction steps. For a training MCR $\mathcal{P}$ with $k$ turns, we extract $(k - l_\text{self} + 1)$ sub-trajectories of the form $\mathcal{P}c = \{(s_1, p_1), \dots, (s_c, p_c), d_l\}$, where $l_\text{self}$ is the number of self-reported symptoms and $c < l$. Each sub-trajectory $\mathcal{P}_c$ serves as a training data.

We calibrate diagnostic confidence through in-batch contrastive learning~\citep{ma2024memorize}, training a lightweight adapter using LoRA~\citep{hu2022lora} to improve the \textit{Diagnosis Agent}’s ability to distinguish among similar diseases. For each patient sub-trajectory $\mathcal{P}c$, the ground-truth diagnosis $d_l$ is treated as the positive instance, while all other candidate diseases serve as negatives. We construct a target distribution $ \text{dist}_\text{target} = [\epsilon, \dots, 1{-}\epsilon, \dots, \epsilon] $, where $\epsilon$ is a label smoothing constant. The \textit{Diagnosis Agent} outputs confidence scores $\{conf_i\}_{i=1}^n$ using the BTP method, yielding a batch-level predictive distribution $\text{dist}_\text{diag}$. The calibration objective minimizes the KL divergence between the target and predicted distributions:
\begin{equation}
\mathcal{L}_{\text{KL}} = \sum_{i=1}^{n} \text{dist}_\text{target}(d_i) \log \frac{\text{dist}_\text{target}(d_i)}{\text{dist}_\text{diag}(d_i)}
\end{equation}

\subsection{Candidate Actions Generation}
Symptom inquiry poses a significant challenge due to the high-dimensional action space, which limits the LLM’s ability to identify the most informative symptoms. A natural solution is to reduce decision complexity by supplying a small set of reliable candidate symptoms. Since each inquiry depends only on the current state, the process satisfies the Markov property, making reinforcement learning (RL) well-suited for this task~\citep{sun2024conversational}. Unlike RLHF approaches that fine-tune LLM parameters—such as GRPO~\citep{ramesh2024group} in DeepSeek-R1~\citep{guo2025deepseek}—we adopt a lightweight RL policy model as an external agent to guide the LLM’s inquiry decisions.

\subsubsection{Observation State and Action Space}
In reinforcement learning, the observation state encodes the information available to the agent at each decision step, while the action space defines the set of allowable actions.

We define the observation state as $\mathcal{S} = [p, c]$, where $p \in \{-1, 0, 1\}^m$ is an $m$-dimensional symptom vector indicating absence ($-1$), unknown status ($0$), or presence ($1$) of each symptom (initialized to $0$), and $c \in \mathbb{R}^n$ is a diagnostic confidence vector over $n$ candidate diseases.

The action space $\mathcal{A} = \{a_i\}_{i=1}^{m+1}$ comprises $m$ inquiry actions—each $a_i$ corresponds to check for the $i$-th symptom—and a termination action $a_{m+1}$ to end the consultation. To reflect clinical heuristics where physicians prioritize symptoms relevant to likely diagnoses~\citep{stanley2013logic}, we introduce a binary action mask $\mathcal{M} \in \{0,1\}^{m+1}$ to constrain the action space. The mask enables actions ($\mathcal{M}_i = 1$) associated with symptoms relevant to the top-$w$ ranked diseases and disables actions that have already selected or deemed irrelevant ($\mathcal{M}i = 0$). The final masked action space is:
\begin{equation} \mathcal{A}_{\text{masked}} = \mathcal{A} \odot \mathcal{M} \end{equation}

\subsubsection{RL Policy Learning}
We adopt an actor-critic architecture to jointly learn the policy $\pi$, which is implemented via multi-layer perceptron (MLP) layers. The policy $\pi$ outputs a log-probability distribution over actions. Training is conducted using Proximal Policy Optimization (PPO)~\citep{schulman2017proximal}, which maximizes the total reward return $\mathcal{R}$, composed of both short-term and long-term components.

The short-term reward $\mathcal{R}_{\text{short}}$ is computed after each doctor-patient interaction:
\begin{equation}
\mathcal{R}_{\text{short}}(S_t, a_t, S_{t+1}) = freq(a_t) + r_{\text{hit}} + r_{\text{rank}},
\end{equation}
where $freq(a_t)$ denotes the frequency of symptom $s_t$ (corresponding to action $a_t$) among the relevant symptoms of the ground-truth disease $d_l$, with negative values assigned to irrelevant symptoms. The term $r_{\text{hit}}$ is positive if $s_t$ is present in the patient profile $\mathcal{P}$ and negative otherwise. The term $r_{\text{rank}}$ measures the change in the confidence ranking of $d_l$ from state $S_t$ to $S_{t+1}$, assigning positive reward for improved ranking, negative for worsened ranking, and zero if unchanged.

The long-term reward $\mathcal{R}_{\text{long}}(d_p)$ assesses the final diagnostic prediction $d_p$, yielding a positive reward if $d_p = d_l$ and a negative reward otherwise.

\subsubsection{Masked Sampling for Candidate Actions} 
The RL policy model's sampling nature inherently prevents the guarantee of optimal actions. However, by performing multiple sampling iterations, we can leverage this characteristic to provide the LLM with a reliable set of candidate actions, thus avoiding decision-making within a large action space.

Specifically, given the current state $\mathcal{S}$, the \textit{Policy Agent} $ \pi $ samples $ N $ actions from the masked action space to construct a candidate action set $ A_{\text{candi}} $ (Equation~\ref{eq:candidate_actions}). If the sampled candidate action set includes the termination action $ a_{m+1} $, the consultation process is terminated.
\begin{equation}
A_{\text{candi}} = \{ a_i \sim \pi(\mathcal{A}_{\text{masked}} \mid \mathcal{S}) \}_{i=1}^{N}
\label{eq:candidate_actions}
\end{equation}

\subsection{Optimal Inquiry Action Selection} 
Upon receiving the candidate action set $A_{\text{candi}}$, the \textit{Inquiry Agent} selects the most informative inquiry action $a_o$, corresponding to symptom $s_o$, by following a set of predefined reasoning strategies. Specifically, it either: (1) attempts to confirm the most probable diagnosis by prioritizing symptoms that are highly representative of the top-ranked disease, or (2) selects the symptom most relevant to the currently collected diagnostic evidence. If none of the candidate actions meet the selection criteria, the \textit{Inquiry Agent} requests the \textit{Policy Agent} to regenerate a new set of actions. By engaging in step-by-step reasoning, the \textit{Inquiry Agent} ensures that the selected symptom $s_o$ maximizes diagnostic value while maintaining interpretability and transparency throughout the decision-making process.

\subsection{Patient Response Simulation}
The \textit{Patient Agent} responds to each inquiry from the \textit{Inquiry Agent} based on the patient profile $ \mathcal{P} $. If the queried symptom $ s_o $ is explicitly documented, it returns the recorded status $ p_o $. However, since real-world Medical Consultation Records (MCRs) typically contain only a limited subset of symptom annotations, many queried symptoms may be undocumented, leading to ineffective queries and ambiguous responses.

To reduce this uncertainty bias, we leverage the disease label $ d_l $ from the MCR and incorporate clinical knowledge for inference. If the queried symptom is uncommon in the clinical presentation of $ d_l $, the agent infers it is likely absent; if the symptom is strongly associated with $ d_l $, it is inferred to be likely present. Once the presence status $ p_o $ is determined, the shared diagnostic memory is updated as:   $ E = E \cup \{(s_o, p_o)\}. $

In DDO, the multi-agent collaborative consultation proceeds for up to $L$ turns, and terminates either when the turn limit is reached or when the termination action is sampled by the \textit{Policy Agent}. The disease with the highest diagnostic confidence is then selected as the final diagnosis.

\section{Experiments}
\subsection{Baselines}
\subsubsection{Traditional Methods.} 
We compare the proposed DDO framework with two state-of-the-art generation-based methods in the MC task: \textbf{MTDiag}~\citep{hou2023mtdiag}, which independently optimizes symptom inquiry and disease diagnosis, and \textbf{HAIformer}~\citep{zhao2024haiformer}, which leverages human-AI collaboration. We additionally include \textbf{EBAD}~\citep{yan2023efficient} as a reinforcement learning-based baseline.

\subsubsection{LLM-Based Methods} 
We compare the DDO framework with three LLM-based methods in the MC task: \textbf{Uncertainty of Thoughts (UoT)}~\citep{hu2024uncertainty}, \textbf{Chain-of-Diagnosis (CoD)}~\citep{chen2024cod}, and \textbf{Direct Prompting (DP)}. UoT plans future inquiries by computing information gain. CoD employs instruction tuning to teach LLMs transparent diagnostic reasoning. DP relies solely on the inherent capabilities of the LLM, without prompt engineering or instruction tuning.

\subsection{Datasets}
We evaluate the proposed DDO framework and the baseline methods on three real-world medical consultation datasets: DXY~\citep{xu2019end}, collected from online medical dialogues, and GMD~\citep{liu2022my} and CMD~\citep{yan2023efficient}, both derived from electronic medical records (EMRs). Dataset statistics are summarized in Table~\ref{tab:dataset_statistics}. Other dataset details can be found in Appendix~\ref{app:datasets_details}.

\begin{table}[ht]
\centering
\resizebox{0.9\columnwidth}{!}{  
\begin{tabular}{l|ccc}
\hline
Dataset & DXY & GMD & CMD \\
\hline
\# Total MCR Samples   & 527   & 2,374 & 5,200 \\
\# Disease Types       & 5     & 12    & 27    \\
\# Symptom Types       & 41    & 118   & 358   \\
\# Avg. Symptoms       & 4.74  & 5.55  & 17.92 \\
\hline
\end{tabular}
}
\caption{Dataset statistics, including the number of medical consultation records (MCR), disease types, symptom types, and average recorded symptoms per sample.}
\label{tab:dataset_statistics}
\end{table}

\begin{table*}
  \centering
  \resizebox{\textwidth}{!}{
  \begin{tabular}{l|c|ccc|ccc|ccc}
    \hline
    \multirow{2}{*}{Method} & \multirow{2}{*}{LLM} & \multicolumn{3}{c|}{DXY} & \multicolumn{3}{c|}{GMD} & \multicolumn{3}{c}{CMD}\\
    \cline{3-11}
    & & $\text{Acc}_\text{init}$ & Acc & Avg.n & $\text{Acc}_\text{init}$ & Acc & Avg.n & $\text{Acc}_\text{init}$ & Acc & Avg.n \\
    \hline     
    \rowcolor{gray!13} \multicolumn{11}{l}{\textbf{Traditional Methods}}\\
    EBAD~\citep{yan2023efficient} & - & - & 72.1 & 7.0 & - & 78.7 & 7.4 & - & 64.1 & 9.0 \\
    MTDiag~\citep{hou2023mtdiag} & - & - & 81.9 & 9.6 & - & \underline{85.9} & 9.6 & - & - & - \\
    HAIformer~\citep{zhao2024haiformer} & - & - & \underline{88.5} & 1.8 & - & \textbf{90.4} & 1.9 & - & \textbf{71.1} & 3.6 \\
    \hline
    \rowcolor{gray!13} \multicolumn{11}{l}{\textbf{LLM-based Methods}}\\
    \multirow{4}{*}{DP} & Qwen2.5-72B-Instruct & 59.6 & 64.4 & 9.0 & 59.8 & 64.4 & 9.9 & 44.2 & 46.8 & 9.8 \\
    & GPT-4o-mini & 57.7 & 61.5 & 10.0 & 57.3 & 65.7 & 10.0 & 45.9 & 49.5 & 10.0 \\
    & Qwen2.5-14B-Instruct & 54.8 & 53.8 & 10.0 & 55.2 & 61.1 & 10.0 & 42.2 & 45.7 & 10.0 \\
    & Qwen2.5-7B-Instruct & 59.6 & 63.5 & 10.0 & 54.8 & 57.3 & 10.0 & 46.8 & 46.2 & 10.0 \\
    \hline
    \multirow{4}{*}{UoT~\citep{hu2024uncertainty}} & Qwen2.5-72B-Instruct & - & 67.3 & 0.1 & - & 68.6 & 0.1 & - & 34.6 & 0.1 \\
    & GPT-4o-mini & - & 64.4 & 0.1 & - & 65.3 & 0.4 & - & 23.0 & 1.3 \\
    & Qwen2.5-14B-Instruct & - & 60.6 & 0.2 & - & 61.1 & 0.6 & - & 32.6 & 1.3 \\
    & Qwen2.5-7B-Instruct & - & 61.5 & 2.2 & - & 71.1 & 0.5 & - & 32.5 & 1.6 \\
    \hline
    \multirow{2}{*}{CoD~\citep{chen2024cod}} & DiagnosisGPT-34B & 61.5 & 53.8 & 3.9 & 54.0 & 44.8 & 3.6 & 46.1 & 34.6 & 3.7 \\
    & DiagnosisGPT-6B & 61.5 & 36.9 & 5.1 & 56.1 & 37.2 & 3.9 & 46.6 & 28.5 & 3.4 \\
    \hline
    \multirow{2}{*}{DDO(Ours)} & Qwen2.5-14B-Instruct & \textbf{66.3} & \textbf{94.2} & 10.0 & \underline{67.8} & 80.3 & 9.8 & \textbf{65.3} & \underline{68.6} & 10.0 \\
    & Qwen2.5-7B-Instruct & \underline{66.3} & 87.5 & 9.9 & \textbf{69.5} & 79.5 & 9.6 & \underline{60.6} & 63.6 & 10.0 \\
    \hline                          
  \end{tabular}}
  \caption{\label{tab:overall_performance}
    Overall performance of DDO and baseline methods on three public medical consultation (MC) datasets. $\text{Acc}_\text{init}$ denotes diagnostic accuracy without any symptom inquiry. Bold numbers indicate the best performance, underlined numbers indicate the second-best. All diagnostic accuracy results are reported as percentages. 
  }
\end{table*}

\subsection{Evaluation Metrics}

\noindent \textbf{Diagnostic Accuracy.}
We use diagnostic accuracy (Acc) as an evaluation metric, which measures the proportion of test cases where the ground-truth disease is correctly identified from a limited set of candidate diseases. This metric reflects the model’s ability to perform accurate differential diagnosis.

\noindent \textbf{Average Turns.} 
The average number of inquiry turns Avg.n is calculated as the total number of symptom queries divided by the number of test samples, reflecting the efficiency and informativeness of the multi-turn consultation process.

\subsection{Implemention Details}
For traditional baselines, we reproduce EBAD and HAIformer following their original implementations, while the MTDiag results are adopted from their paper. All LLM-based baselines are re-implemented, where UoT adopts its pruned version to improve planning efficiency. The backbone LLMs include the Qwen2.5 series~\citep{yang2024qwen2}, GPT-4o-mini~\citep{achiam2023gpt}, and DiagnosisGPT~\citep{chen2024cod}, which is fine-tuned to align with the CoD consultation process. For methods that involve parameter tuning, including CoD and our proposed DDO, we use locally deployed LLMs; other LLM-based baselines rely on API-based models. The pruning percentage in UoT is set to -0.5. The threshold $\tau $ in CoD is set to 0.5. All LLM-based approaches adopt the same LLM for both doctor and patient agents. The maximum of doctor-patient interaction turns $L$ is set to 10 for all the methods. More implementation details can be found in the Appendix~\ref{app:implementation_details}.

\subsection{Overall Performance}
Table~\ref{tab:overall_performance} summarizes the main experimental results of the proposed DDO framework and baseline methods across the three MC datasets.

\noindent \textbf{Comparison with Traditional Methods.}
DDO achieves diagnostic accuracy on par with traditional baselines while substantially reducing training overhead. For instance, the SOTA baseline HAIformer adopts a multi-stage training pipeline requiring hundreds of epochs for training its diagnostic module, while DDO only needs a few epochs for confidence calibration—less than one epoch on both the GMD and CMD datasets. This efficiency stems from the strong generalization capabilities of LLMs, which enable effective domain adaptation with a small number of parameter tuning. Moreover, the inherent reasoning ability of LLMs contributes to the interpretability of the MC task.

\noindent \textbf{Comparison with LLM-based Methods.}
Compared to other LLM-based methods, DDO significantly improves diagnostic effectiveness. After symptom inquiry, it boosts diagnostic accuracy by an average of 24.6\%, 11.3\%, and 3.2\% on the DXY, GMD, and CMD datasets, respectively, over initial diagnoses based only on self-reported symptoms. DDO consistently achieves the highest accuracy, notably 94.2\% on DXY. The DP baseline reflects the raw inquiry behavior of LLMs, where the lack of external guidance results in arbitrary questioning and unreliable diagnoses. UoT improves upon DP by using LLM-based planning to prioritize symptoms with the highest expected information gain. However, its pruning strategy—eliminating candidate diseases as soon as key symptoms are denied—restricts comprehensive evidence gathering, often resulting in a small Avg.n and ultimately limiting diagnostic performance. CoD attempts to jointly optimize symptom inquiry and diagnosis via large-scale synthetic reasoning data but yields negative performance gains. This is likely due to the intrinsic differences between the two sub-tasks, which hinder effective unified learning. In contrast, DDO decouples the two decision-making processes in the MC task, enhancing them separately through a lightweight RL policy and a diagnostic adapter. This modular approach yields substantial gains in diagnostic accuracy.

\begin{table}
  \centering
  \resizebox{\columnwidth}{!}{
  \begin{tabular}{l|cc|cc|cc}
    \hline
    \multirow{2}{*}{Method} & \multicolumn{2}{c|}{DXY} & \multicolumn{2}{c|}{GMD} & \multicolumn{2}{c}{CMD}\\
    \cline{2-7}
    & Acc & Avg.n & Acc & Avg.n & Acc & Avg.n \\
    \hline     
    DDO(Ours) & \textbf{87.5} & 9.9 & \textbf{79.5} & 9.6 & \textbf{63.6} & 10.0  \\
    \textit{w/o} adapter & \underline{86.5} & 9.9 & \underline{78.7} & 9.6 & 54.2 & 10.0  \\
    \textit{w/o} policy & 77.7 & 10.0 & 73.6 & 10.0 & 60.3 & 10.0  \\
    \textit{w/o} masking & 83.5 & 9.9 & 74.5 & 10.0 & 61.8 & 10.0 \\
    \textit{w/o} retry & 83.5 & 10.0 & 78.7 & 9.8 & 63.0 & 10.0  \\
    \textit{w/o} decision & 84.6 & 9.8 & 78.2 & 9.8 & \underline{63.2} & 10.0  \\
    \hline   
  \end{tabular}}
  \caption{\label{tab:ablation_study}
    Ablation results of DDO. \textit{w/o} adapter denotes removing the diagnostic adapter. \textit{w/o} policy and \textit{w/o} decision use the LLM and RL model for symptom inquiry, respectively. \textit{w/o} masking disables action space masking.  \textit{w/o} retry omits candidate actions regeneration.
  }
\end{table}

\subsection{Ablation Study}
As shown in Table~\ref{tab:ablation_study}, we perform ablation experiments on three MC datasets, using Qwen2.5-7B-Instruct as the LLM backbone.

\noindent \textbf{Impact of Diagnostic Adapter.} 
Removing the diagnostic adapter (\textit{w/o} adapter)—using only the vanilla BTP method to estimate diagnostic confidence—results in a drop in diagnostic accuracy, with the largest decline observed on the CMD dataset. This demonstrates the effectiveness of the in-batch contrastive learning–based adapter in enhancing the disease discrimination ability of LLMs.

\noindent \textbf{Impact of RL-LLM Collaboration.}
To assess the effectiveness of RL-LLM collaborative symptom inquiry process, we conduct ablation experiments in which symptom inquiry is performed solely by the RL policy (\textit{w/o} decision) or solely by the LLM (\textit{w/o} policy). Results show that both variants perform significantly worse, with the LLM-only variant exhibiting a greater performance drop. This highlights the advantage of multi-agent collaboration in DDO for conducting effective symptom inquiry. Moreover, removing action space masking (\textit{w/o} masking) or disabling the regeneration mechanism for low-quality actions (\textit{w/o} retry) also degrades performance, underscoring their role in ensuring reliable inquiry decisions.

\begin{table}
  \centering
  \resizebox{\columnwidth}{!}{
  \begin{tabular}{l|cc|cc|cc}
    \hline
    \multirow{2}{*}{Method} & \multicolumn{2}{c|}{DXY} & \multicolumn{2}{c|}{GMD} & \multicolumn{2}{c}{CMD}\\
    \cline{2-7}
    & $\text{Acc}_\text{init}$ & Acc & $\text{Acc}_\text{init}$ & Acc & $\text{Acc}_\text{init}$ & Acc \\
    \hline     
    Numerical & 62.5 & 78.8 & 53.6 & 74.5 & 29.7 & 39.4 \\
    Numerical-SC & \underline{65.4} & 77.9 & 54.4 & 74.9 & 32.8 & 45.3 \\
    FirstLogit & 59.6 & 70.2 & 60.3 & 74.1 & 40.5 & 43.7 \\
    AvgLogit & 42.3 & 75.0 & 59.4 & 74.5 & 25.2 & 28.8 \\
    BTP & 63.5 & \underline{86.5} & \underline{63.6} & \underline{78.7} & \underline{54.2} & \underline{54.2} \\
    BTP-adapter(Ours) & \textbf{66.3} & \textbf{87.5} & \textbf{69.5} & \textbf{79.5} & \textbf{60.6} & \textbf{63.6} \\
    \hline   
  \end{tabular}}
  \caption{\label{tab:diagnosis_comparison}
    Comparison of diagnostic performance across different confidence estimation methods for LLMs.
  }
\end{table}

\subsection{Evaluation of Diagnostic Performance in LLM Confidence Estimation}

To further assess the diagnostic effectiveness of our BTP-adapter, we compare it with several alternative confidence estimation methods for LLMs using three MC datasets. All methods are evaluated with the same initial and final symptom sequences, and the results are presented in Table~\ref{tab:diagnosis_comparison}.

\noindent \textbf{Decoding-based Methods.}
Numerical and Numerical-SC prompt the LLM to directly generate a confidence score between 0 and 1~\citep{li2024mediq}, with SC indicating the use of Self-Consistency~\citep{wang2022self}. These approaches show significantly lower diagnostic accuracy than the BTP-adapter, highlighting the limitations of decoding-based confidence estimation and the advantages of our logits-based strategy.

\noindent \textbf{Logits-based Methods.}
Similar to our approach, FirstLogits and AvgLogits~\citep{ma2025inferring} estimate confidence based on the logits of the first generated token. However, their diagnostic performance is notably inferior to that of the BTP-adapter. This may be due to their practice of computing confidence scores for all candidate diseases in a single generation process, which can lead to context interference. In contrast, BTP-adapter independently evaluates each candidate diagnosis, effectively mitigating such interference. We also compare against the original BTP method without the diagnostic adapter. Incorporating the adapter consistently enhances diagnostic accuracy, particularly in the initial diagnosis, which is critical for guiding effective symptom inquiry during the early stages of MC.

\begin{figure}[t]
    \centering
    \includegraphics[width=\columnwidth]{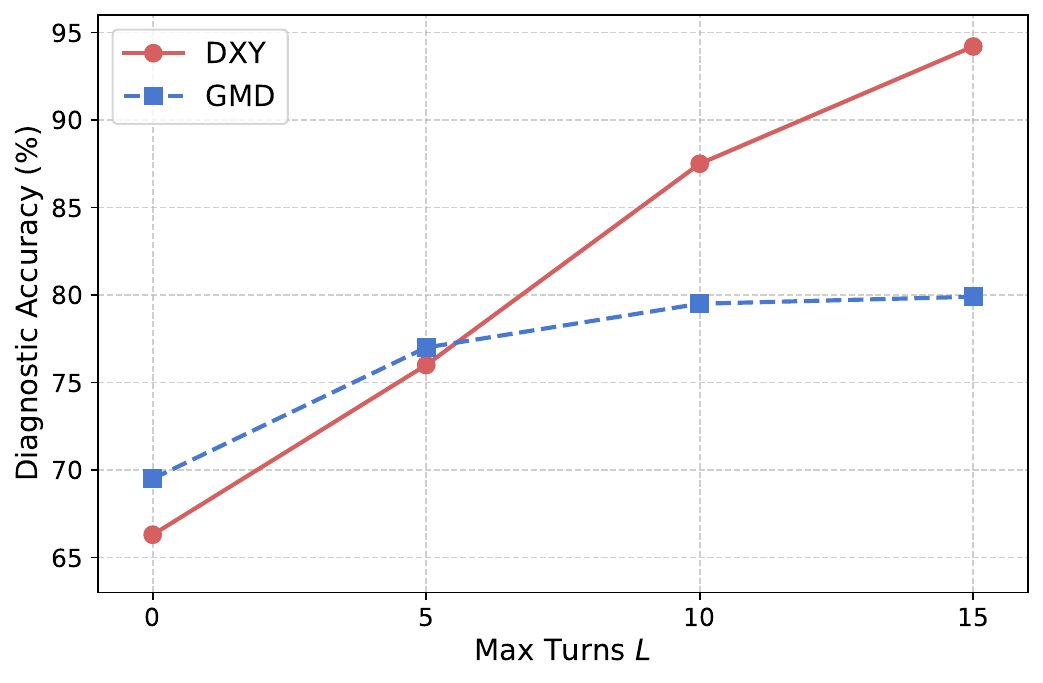}
    \caption{Effect of max turns $L$.}
    \label{fig:max_turns_effect}
\end{figure}

\begin{figure}[t]
    \centering
    \includegraphics[width=\columnwidth]{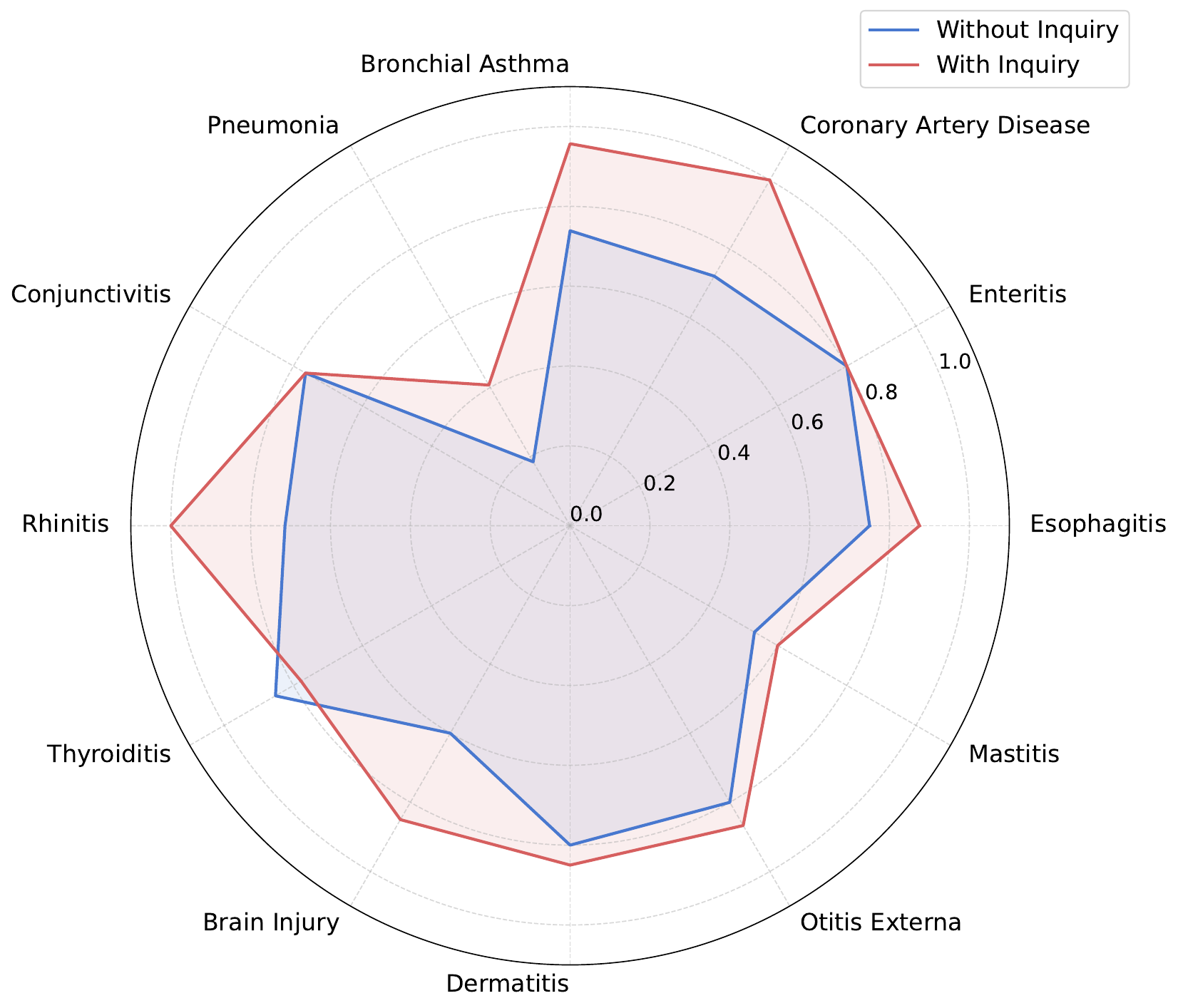}
    \caption{Diagnosis performance at the disease level on the GMD dataset.}
    \label{fig:different_diseases_radar}
\end{figure}

\subsection{Effect of Max Turns \texorpdfstring{$L$}{L}}
As shown in Figure~\ref{fig:max_turns_effect}, we evaluate the diagnostic performance of the DDO framework on the DXY and GMD datasets for different maximum interaction turns ($L = 0/5/10/15$). The line charts show an upward trend, indicating that increasing the maximum number of turns $L$ generally improves diagnostic accuracy. This suggests that the symptom inquiry process in DDO effectively collects critical diagnostic evidence. Notably, the most significant improvements occur in the early stages, with the gains diminishing as more turns are added—especially evident on the GMD dataset. One possible explanation is that DDO prioritizes inquiries for diseases with high initial diagnostic confidence. When the ground-truth disease $d_l$ initially ranks lower, additional inquiries may offer diminishing returns in terms of diagnostic accuracy.

\subsection{Diagnostic Effectiveness Across Different Diseases}
\label{sec:diagnosis_accross_diseases}

To assess diagnostic performance of DDO at a fine-grained level, we visualize the diagnostic accuracy before and after symptom inquiry across 12 candidate diseases in the GMD dataset, as shown in Figure~\ref{fig:different_diseases_radar}. The results indicate that multi-turn symptom inquiry substantially enhances diagnostic accuracy for most diseases. Notably, the final accuracy for \textit{Coronary Artery Disease} and \textit{Rhinitis} reaches 100\%, highlighting the effectiveness of the collaborative symptom collection between the \textit{Policy Agent} and \textit{Inquiry Agent} in DDO. However, for certain diseases such as \textit{Conjunctivitis} and \textit{Thyroiditis}, the inquiry process yields no improvement in diagnostic accuracy, and even shows a slight decline for \textit{Thyroiditis}. Further analysis reveals that some MCRs for these diseases include only one or two vague or non-specific self-reported symptoms, which causes these diseases to be ranked relatively low initially, making it difficult for the Agents to identify critical symptoms for accurate prediction.

In addition, Table~\ref{tab:f1} reports both macro-F1 and micro-F1 scores when comparing DDO with the strongest baseline HAIformer. Although DDO does not surpass this baseline, it achieves comparable performance within fewer training epochs. The micro-F1 results demonstrate that DDO preserves overall diagnostic accuracy, while the macro-F1 results highlight that DDO maintains relatively balanced diagnostic effectiveness across different diseases, avoiding bias toward high-frequency classes.

\begin{table}
  \centering
  \resizebox{\columnwidth}{!}{
  \begin{tabular}{l|cc|cc|cc}
    \hline
    \multirow{2}{*}{Method} & \multicolumn{2}{c|}{DXY} & \multicolumn{2}{c|}{GMD} & \multicolumn{2}{c}{CMD}\\
    \cline{2-7}
    & macro-F1 & micro-F1 & macro-F1 & micro-F1 & macro-F1 & micro-F1 \\
    \hline     
    HAIformer & 88.8 & 88.5 & \textbf{90.1} & \textbf{90.4} & \textbf{69.3} & \textbf{71.1} \\
    DDO(Ours) & \textbf{93.6} & \textbf{93.3} & 81.6 & 80.3 & 65.8 & 67.4 \\
    \hline   
  \end{tabular}}
  \caption{\label{tab:f1}
    Macro-F1 and Micro-F1 scores of DDO and the SOTA baseline on three MC datasets.
  }
\end{table}

\section{Discussion}

\noindent \textbf{Impact of Action Space on RL Policy Learning.}
In general, the larger the action space of an RL policy model, the more challenging the decision-making process becomes, and the more difficult the optimization tends to be. For example, we observed that with a smaller action space size (42 for DXY), the RL policy was able to sample more informative symptoms, whereas with a larger action space size (359 for CMD), the policy was more prone to sampling irrelevant ones. This motivates our use of masked sampling and LLM-voting during inference: by pruning the action space and leveraging RL–LLM collaboration, we aim to reduce the decision burden placed on the RL policy.

\noindent \textbf{Improving Diagnostic Gains in Challenging Diseases.} 
For certain diseases, diagnostic accuracy exhibits only modest gains—or even slight declines—after symptom inquiry (e.g., Thyroiditis in GMD). As discussed in Section~\ref{fig:different_diseases_radar}, this likely stems from insufficient information in the initial symptom presentation of some MCR cases, which constrains the Agents’ ability to identify and query informative symptoms. Possible remedies include: (1) enhancing the initial Top-K accuracy of the Diagnosis Agent for these challenging diseases, since a stronger initial ranking increases the chance of retrieving relevant symptoms and thereby improves inquiry effectiveness; and (2) refining the inquiry strategy by leveraging the LLM’s reasoning capacity to filter out irrelevant diseases, thus avoiding excessive focus on unrelated conditions.

\noindent \textbf{Joint Optimization of the Diagnosis Adapter and RL Policy.}  
In this work, we assign different optimization objectives to the diagnosis module and the symptom inquiry module, and adopt a cascaded optimization scheme. This dual-decision optimization effectively enhances the overall performance of LLM-based medical consultation. Nevertheless, LLM-RL cascading training can be inherently unstable. We believe that further improvements could be achieved by jointly optimizing the diagnosis adapter and the RL policy model. Such joint training may not only enable the diagnosis adapter to acquire more robust multi-disease diagnostic capabilities better aligned with interactive doctor–patient scenarios, but also facilitate more stable reward learning for the RL policy.

\noindent \textbf{Mitigating Inference Overhead from Multi-Agent Collaboration.}  
Our DDO framework employs four collaborating agents to accomplish the MC task, substantially enhancing the diagnostic performance of LLM-based methods. However, this multi-agent design also introduces significant inference overhead—for instance, a single multi-turn consultation per MCR typically takes 3–5 minutes. To mitigate this cost, a promising direction inspired by \citep{chopra2025feedback} is to store completed consultation trajectories in a memory tree. When processing a new MCR, the system can first retrieve similar cases from the tree; if a close match is identified, the consultation can follow the retrieved trajectory, thereby reducing the number of LLM inference calls.

\section{Conclusion}
In this paper, we propose DDO, a novel LLM-based multi-agent collaborative framework designed to address the mismatch between existing LLM-based methods and the dual-decision nature of medical consultation (MC), which involves both sequential symptom inquiry and diagnosis over a constrained set of candidate diseases. DDO decouples these two decision-making processes and optimizes them with distinct objectives: it improves disease discrimination through a plug-and-play diagnostic adapter, and enhances information gathering via the synergy of an reinforcement learning-based policy agent an LLM-based inquiry agent. Experiments on three public MC datasets show that DDO consistently outperforms existing LLM-based baselines and achieves performance competitive with state-of-the-art generation-based methods, demonstrating its effectiveness in the MC task.

\section*{Limitations}
While we propose DDO to enhance the effectiveness of LLMs in the medical consultation task, several limitations remain: \textbf{(1) Label granularity}: During confidence calibration, DDO assigns a target confidence of 1 to the ground-truth disease and a small constant to all others. This hard-labeling scheme may hinder the model's ability to softly distinguish between clinically similar diseases. \textbf{(2) Model deployment:} Since DDO requires training a diagnostic adapter for the underlying LLM, it is currently incompatible with API-based LLMs and must be deployed with locally hosted models. \textbf{(3) Inference efficiency:} DDO involves multi-agent reasoning, which introduces inference latency compared to traditional deep learning methods.

\section*{Ethical Consideration}
Due to the hallucination problem inherent in large language models, they may generate content that is not factually accurate. Therefore, the DDO framework proposed in this paper is intended solely for academic research. In real-world scenarios, medical decisions should always be based on professional diagnoses made by qualified physicians.

\section*{Acknowledgments}
This work was supported in part by National Natural Science Foundation of China (No. 62576363), the Natural Science Foundation of Hunan Province (No. 2025JJ50342) and Changsha Municipal Natural Science Foundation Project (No. kq2502144). This work was carried out in part using computing resources at the High-Performance Computing Center of Central South University.

\bibliography{custom}

\appendix

\section{Other Implementation Details}
\label{app:implementation_details}

\subsection{Datasets Details}
\label{app:datasets_details}
We use three public medical consultation (MC) datasets: DXY\footnote{\url{https://github.com/HCPLab-SYSU/Medical_DS}}~\citep{xu2019end}
, which is the most widely used online dialogue–based MC dataset in prior work, and GMD\footnote{\url{https://github.com/lwgkzl/BR-Agent}}~\citep{liu2022my}
 together with CMD\footnote{\url{https://github.com/YanPioneer/EBAD}}~\citep{yan2023efficient}
, which are representative of electronic medical record–based MC datasets. These datasets were released for academic research and have been de-identified by their original authors. We reviewed the accompanying papers and code repositories for licensing information: GMD is explicitly provide an MIT license, while DXY and CMD, though lacking formal licenses, clearly state in their papers that the data is intended for public research use. In our study, we strictly follow these terms and use the datasets solely for research purposes. All three datasets are primarily in Chinese, with GMD additionally offering an English version.

For validation, we apply stratified sampling to the training sets of DXY and CMD to create development sets (GMD has already included a predefined split). The final train/dev/test splits for DXY, GMD, and CMD are 318/103/103, 1912/239/239, and 3379/671/1342, respectively.

Since MCRs in DXY and GMD contain relatively few symptoms on average—only 4.74 and 5.55 per record, respectively—this sparsity may hinder the reliable calibration of diagnostic confidence in LLMs. Therefore, when constructing the confidence calibration dataset, we also provide an augmented version of the data to address this issue, enriching MCRs with additional symptoms to enhance the adapter's long-range diagnostic capability. Specifically, for each training MCR with a small number of symptoms, we sample additional implicit symptoms based on disease knowledge extracted from the training data. As a result, the calibration training dataset contains 2,185 instances for DXY (originally 553), 13,598 for GMD (originally 4,837), and 54,608 for CMD. 

To improve training efficiency, we do not include all diseases in each contrastive batch when constructing confidence calibration data. Instead, each ground-truth disease is paired with four clinically similar candidate diseases for comparison.

\subsection{Training Details}
\label{app:training_details} 
Hyperparameters and model checkpoints were selected based on the validation dataset, considering both Top-K accuracy and diagnostic effectiveness across diseases. The key configurations are summarized in Table~\ref{tab:calibration_hyperparameters}, Table~\ref{tab:rl_hyperparameters} and Table~\ref{tab:llm_rl_hyperparameters}. 

During  confidence calibration, for Qwen2.5-7B-Instruct, several promising checkpoints were retained as candidate adapters for the subsequent RL training, while for other LLMs only the best-performing checkpoint was used for inference.

LLM-RL cascading training can be inherently unstable. To mitigate this, each candidate adapter was employed to train a policy model. The resulting models were then evaluated, and the one achieving the highest Top-K accuracy was selected as the final Policy Agent. This agent was subsequently deployed across different LLM backbones, demonstrating both robustness and transferability.

\begin{table}[ht]
\centering
\resizebox{0.8\columnwidth}{!}{
\begin{tabular}{c|c|c|c}
\hline
\textbf{Hyperparameters} & \textbf{DXY} & \textbf{GMD} & \textbf{CMD} \\
\hline
max training epochs & 5 & 1 & 1 \\
global batch size & 8 & 8 & 8 \\
learning rate & 5e-5 & 5e-5 & 5e-5 \\
lora rank & 16 & 16 & 16 \\
in-bacth group length & 5 & 5 & 5 \\
\hline
\end{tabular}
}
\caption{Hyperparameters for confidence calibration.}
\label{tab:calibration_hyperparameters}
\end{table}

\begin{table}[ht]
\centering
\resizebox{\columnwidth}{!}{
\begin{tabular}{c|c|c|c}
\hline
\textbf{Hyperparameters} & \textbf{DXY} & \textbf{GMD} & \textbf{CMD} \\
\hline
actor hidden layers & [256,128,128] & [256,128,128] & [512,256,256] \\
critic hidden layers & [64] & [64] & [128] \\
learning rate & 5e-5 & 5e-5 & 5e-5 \\
batch size & 64 & 128 & 128 \\
steps per update & 1024 & 2048 & 2048 \\
epochs & 5 & 5 & 5 \\
total steps & 51200 & 102400 & 102400 \\
hitting reward & 0.5 & 0.5 & 0.5 \\
ranking reward & 0.5 & 0.5 & 0.5 \\
diagnosis reward & 1.0 & 1.0 & 1.0 \\
frequency penaty & 0.2 & 0.2 & 0.2 \\
window size & 3 & 5 & 5 \\
floor turns & 3 & 6 & 5 \\
\hline
\end{tabular}
}
\caption{Hyperparameters for RL policy model's training and inference.}
\label{tab:rl_hyperparameters}
\end{table}

\begin{table}[ht]
\centering
\resizebox{0.8\columnwidth}{!}{
\begin{tabular}{c|c|c|c}
\hline
\textbf{Hyperparameters} & \textbf{DXY} & \textbf{GMD} & \textbf{CMD} \\
\hline
sampling times & 6 & 6 & 7 \\
window size & 3 & 4 & 5 \\
floor turns & 3 & 5 & 5 \\
retry times & 1 & 2 & 2 \\
\hline
\end{tabular}
}
\caption{Hyperparameters for LLM–RL collaboration in symptom inquiry.}
\label{tab:llm_rl_hyperparameters}
\end{table}

\subsection{Model Deployment}
\label{app:model_deployment}
We locally deployed the LLMs used in our DDO framework on GPU devices. Specifically, Qwen2.5-7B-Instruct was run on an NVIDIA RTX 3090 GPU, while Qwen2.5-14B-Instruct was run on an NVIDIA Tesla V100 GPU. For reproducing the Chain-of-Diagnosis baseline~\citep{chen2024cod}, we deployed DiagnosisGPT-6B and DiagnosisGPT-34B using 1 and 3 NVIDIA Tesla V100 GPUs, respectively. For reproducing EBAD~\citep{yan2023efficient} and HAIformer~\citep{zhao2024haiformer}, we used a single NVIDIA RTX 3090 GPU. For implementing the Direct Prompting baseline and reproducing the Uncertainty of Thoughts baseline~\citep{hu2024uncertainty}, we utilized the Qwen2.5-Instruct API provided by the Siliconflow platform\footnote{\url{https://www.siliconflow.cn/}} and the ChatGPT API provided by the ChatAnywhere platform\footnote{\url{https://chatanywhere.apifox.cn/}}. 

Based on Stable-Baselines3\footnote{\url{https://github.com/DLR-RM/stable-baselines3}}, the reinforcement learning policy model in DDO is trained and deployed on a single NVIDIA RTX 3090 GPU.

\section{Statistical Results}
\label{app:detailed_experiments}

\subsection{Standard Errors and Confidence Intervals}
\label{app:standard_errors_and_confidence_intervals}
We report the standard errors and confidence intervals of the results for the proposed DDO framework on the three MC datasets in Table~\ref{tab:se_ci}, based on experiments conducted with different random seeds. For efficiency considerations, we use Qwen2.5-7B-Instruct as the LLM backbone. The comparison between our proposed DDO framework and the baseline methods in the main experiments was performed using the random seed that achieved the best overall performance.

The statistical results in Table~\ref{tab:se_ci} indicate that the diagnostic performance of DDO is relatively stable across different random seeds. The standard errors are very small (below 1.0 on all datasets), and the 95\% confidence intervals are also narrow (e.g., only 0.6 on GMD), suggesting that repeated runs of DDO yield relatively consistent results.

\begin{table}[ht]
\centering
\resizebox{\columnwidth}{!}{  
\begin{tabular}{l|ccc}
\hline
\textbf{Dataset} & \textbf{Mean Accuracy} & \textbf{Standard Error} & \textbf{95\% CI (\%)} \\
\hline
DXY & 84.4 & 0.8 & 2.3 \\
GMD & 79.6 & 0.1 & 0.6 \\
CMD & 63.1 & 0.4 & 1.6 \\
\hline
\end{tabular}
}
\caption{Mean accuracy, standard error (SE), and 95\% confidence interval (CI) of DDO for final diagnostic accuracy across three MC datasets.}
\label{tab:se_ci}
\end{table}

\subsection{Significance Testing}

We conducted paired significance tests between the diagnostic performance of our proposed DDO framework and the state-of-the-art baseline HAIformer on the same set of random seeds. On the DXY dataset, DDO demonstrates a statistically significant improvement over HAIformer ($p<0.05$). However, on the GMD and CMD datasets, DDO achieves lower diagnostic accuracy than HAIformer, and the differences are statistically significant ($p<0.01$). These results highlight potential challenges in scaling LLM-based medical consultation methods to datasets with more diverse disease categories.

\begin{table}[ht]
\centering
\resizebox{0.9\columnwidth}{!}{%
\begin{tabular}{l|ccc}
\hline
\textbf{Metric} & \textbf{DXY} & \textbf{GMD} & \textbf{CMD} \\
\hline
Paired t-test p-value & 0.0355 & 0.0002 & 0.0046 \\
\hline
\end{tabular}%
}
\caption{Statistical significance of DDO compared to HAIformer across three MC datasets.}
\label{tab:significance_testing}
\end{table}

\section{Case Study}
\label{app:case_study}
Table~\ref{tab:case_study} shows a medical consultation case. The patient initially reported the symptom \textit{runny nose}, and DDO used this information to conduct multiple rounds of symptom inquiries to gather more evidence. In each round, DDO first provides a set of candidate inquiry actions via the \textit{Policy Agent}. The \textit{Inquiry Agent} then selects the most appropriate action based on reasoning. The \textit{Patient Agent} responds with the presence or absence of symptoms based on the Medical Consultation Record (MCR). For symptoms not recorded in the MCR, the \textit{Patient Agent} infers their likely presence or absence based on the clinical presentation of the disease. The \textit{Diagnosis Agent} evaluates the diagnostic confidence for each candidate disease. In this case, after reaching the maximum number of interaction rounds, the disease with the highest diagnostic confidence—\textit{allergic rhinitis (AR)}—was identified as the patient's final diagnosis, which aligns with the ground-truth diagnosis in the MCR.

\section{Prompt Templates}
\label{app:prompts}

Tables~\ref{tab:demo_disease_diagnosis_prompt}-\ref{tab:demo_patient_simulation_prompt} show the prompts used in our DDO framework. Tables~\ref{tab:dp_interaction_decision_prompt}-\ref{tab:dp_disease_diagnosis_prompt} show the prompts used in the DP baseline.

\onecolumn
{\small
\begin{longtable}{p{0.96\textwidth}}
  \hline
  \textbf{Self Report (Explicit Symptoms):} The patient has symptoms of cough, restlessness, and runny nose. \\
  \textbf{Explicit Symptoms \& Implicit Symptoms:} \{"cough": True, "restlessness": True, "runny nose": True, "anorexia": True, "short of breath": True\} \\
  \textbf{Disease Label:} pneumonia \\
  \textbf{Candidate Diseases:} ["hand-foot-and mouth disease (HFMD)", "upper respiratory tract infection (URTI)", "pneumonia", "allergic rhinitis (AR)", "diarrhea"] \\
  \textbf{Diagnostic Confidence of Top Diseases:} \{"URTI": 0.68, "pneumonia": 0.16, "AR": 0.16\} \\ 
  \hline
    
    \rowcolor{gray!15}
    \multicolumn{1}{l}{\textbf{Turn 1}} \\
    \textbf{Candidate Actions:} ["check for fever", "check for vomiting", "check for short of breath", "check for pharyngeal discomfort"] \\
    \textbf{Selected Symptom for Inquiry:} fever \\
    \textbf{Selection Reasoning:} First, we examine the current diagnostic confidence scores: upper respiratory infection (0.68), pneumonia (0.16), and allergic rhinitis (0.16). Since upper respiratory infection has a significantly higher confidence, we prioritize symptoms associated with it. Looking at symptom frequencies for this disease, fever (0.507) is the most common, followed by vomiting (0.233), throat discomfort (0.164), and short of breath (0.123). Based on Strategy 1, we should ask the patient about fever next, as it is the most indicative symptom. \\
    \textbf{Symptom Status:} True \\
    \textbf{Response Reasoning:} To assess whether the patient is likely to have the symptom "fever," we examine its frequency relative to other listed symptoms. A higher frequency—i.e., a higher rank in the sorted list—indicates the symptom is more typical and thus more likely to be present in the patient. According to the information, "fever" has a frequency of 0.533. When we sort all symptoms by frequency in descending order, "fever" ranks 3rd, following "cough" (0.883) and "sputum production" (0.65). This places "fever" in the high-frequency range. Therefore, based on this analysis, we determine the likelihood of the patient having the symptom "fever" to be high, and the final judgment is True. \\
    \textbf{Diagnostic Confidence of Top Diseases:} \{"URTI": 0.45, "pneumonia": 0.45, "HFMD": 0.11\} \\
    \hline

    \rowcolor{gray!15}
    \multicolumn{1}{l}{\textbf{Turn 2}} \\
    \textbf{Candidate Actions:} ["check for herpes", "check for sneezing", "check for pharyngeal discomfort", "check for short of breath", "check for vomiting"] \\
    \textbf{Selected Symptom for Inquiry:} short of breath \\
    \textbf{Selection Reasoning:} The current diagnostic confidence scores are: upper respiratory infection (0.45), pneumonia (0.45), and hand, foot, and mouth disease (0.11). Since no single disease stands out, we can’t apply Strategy 1. So we use Strategy 2, which selects symptoms related to the patient’s current presentation—cough, irritability, runny nose, and fever. Among the candidate symptoms (herpes, sneezing, throat discomfort, short of breath, vomiting), short of breath is the most relevant, as it is strongly associated with pneumonia. Therefore, we choose short of breath as the next symptom to inquiry about. \\
    \textbf{Symptom Status:} True \\
    \textbf{Response Reasoning:} Short of breath is recorded in MCR, which indicates that the patient has this symptom. \\
    \textbf{Diagnostic Confidence of Top Diseases:} \{"pneumonia": 0.68, "URTI": 0.16, "HFMD": 0.16\} \\
    \hline
    
    \rowcolor{gray!15}
    \multicolumn{1}{l}{\textbf{Turn 3}} \\
    \textbf{Candidate Actions:} ["check for pulmonary moist rales", "check for bubble-like spitting", "check for sneeze", "check for vomiting"] \\
    \textbf{Selected Symptom for Inquiry:} pulmonary moist rales \\
    \textbf{Selection Reasoning:} The current diagnostic confidence scores are: pneumonia (0.68), upper respiratory infection (0.16), and hand, foot, and mouth disease (0.16). Since pneumonia stands out with the highest confidence, we prioritize its typical symptoms. Among the candidate symptoms—pulmonary moist rales (0.167), bubble blowing (0.167), sneezing (0.117), and vomiting (0.383)—both pulmonary moist rales and bubble blowing are more specific to pneumonia, despite vomiting having a higher frequency. Following Strategy 1, we choose pulmonary moist rales as the next symptom to ask the patient. \\
    \textbf{Symptom Status:} False \\
    \textbf{Response Reasoning:} To assess whether the patient is likely to have the symptom pulmonary moist rales, we first examine its frequency in pneumonia cases and compare it with other symptoms. The frequency of pulmonary moist rales is 0.167. When we sort all listed symptoms by their frequency in pneumonia, pulmonary moist rales ranks relatively low—below common symptoms like cough (0.883), sputum production (0.65), and fever (0.533), and only slightly higher than symptoms like fatigue, loose stools, and sweating. This suggests that pulmonary moist rales is not among the most typical symptoms of pneumonia. Therefore, we conclude that the patient is less likely to have this symptom, and the judgment is False. \\
    \textbf{Diagnostic Confidence of Top Diseases:} \{"pneumonia": 0.68, "URTI": 0.16, "diarrhea": 0.16\} \\
    \hline
    
    \rowcolor{gray!15}
    \multicolumn{1}{l}{\textbf{Turn 4}} \\
    \textbf{Candidate Actions:} ["check for anorexia", "check for vomiting", "check for lethargy"] \\
    \textbf{Selected Symptom for Inquiry:} vomiting \\
    \textbf{Selection Reasoning:} The current diagnostic confidence scores are: pneumonia (0.68), upper respiratory infection (0.16), and pediatric diarrhea (0.16). Since pneumonia has the highest confidence, we focus on symptoms related to it. Among the candidate symptoms, vomiting has the highest frequency in pneumonia cases at 0.383, compared to anorexia (0.15) and fatigue (0.083). Based on this, we choose vomiting as the next symptom to ask the patient about. \\
    \textbf{Symptom Status:} True \\
    \textbf{Response Reasoning:} To determine whether the patient is likely to have the symptom "vomiting," we examine its frequency among pneumonia patients, which is 0.383. Although this is lower than symptoms like cough (0.883), sputum production (0.65), fever (0.533), runny nose (0.467), and short of breath (0.433), it is higher than others such as nasal obstruction (0.267) and irritability (0.233). This indicates that vomiting is relatively common in pneumonia cases. While it may not be the most typical symptom, its frequency suggests it has moderate representativeness. Therefore, we conclude that the likelihood of the patient having this symptom is relatively high, and the judgment result is True. \\
    \textbf{Diagnostic Confidence of Top Diseases:} \{"pneumonia": 0.68, "URTI": 0.16, "HFMD": 0.16\} \\
    \hline
    
    \rowcolor{gray!15}
    \multicolumn{1}{l}{\textbf{Turn 5}} \\
    \textbf{Candidate Actions:} ["check for rash", "check for expectoration", "check for anorexia", "check for herpes", "check for sneeze"] \\
    \textbf{Selected Symptom for Inquiry:} expectoration \\
    \textbf{Selection Reasoning:} Pneumonia currently has the highest diagnostic confidence (0.68), much higher than upper respiratory infection and hand-foot-mouth disease (both 0.16). So, we focus on symptoms typical of pneumonia. Among the candidate symptoms, only expectoration (coughing up phlegm) has a high occurrence in pneumonia (65\%), while others like rash, anorexia, sneezing, and herpes are either rare or not typical. Therefore, the most appropriate next symptom to ask about is expectoration. \\
    \textbf{Symptom Status:} True \\
    \textbf{Response Reasoning:} To determine whether the patient is likely to have the symptom expectoration, we refer to the symptom frequency data for pneumonia. The frequency of expectoration is 0.65, which is relatively high—only lower than cough (0.883) and fever (0.533). This indicates that expectoration is a common and typical symptom of pneumonia. Therefore, it's reasonable to conclude that the patient is likely to have this symptom. Final judgment: True. \\
    \textbf{Diagnostic Confidence of Top Diseases:} \{"pneumonia": 0.68, "URTI": 0.16, "diarrhea": 0.16\} \\
    \hline
    
    \rowcolor{gray!15}
    \multicolumn{1}{l}{\textbf{Turn 6}} \\
    \textbf{Candidate Actions:} ["check for lethargy", "check for sneeze", "check for nasal obstruction"] \\
    \textbf{Selected Symptom for Inquiry:} nasal obstruction \\
    \textbf{Selection Reasoning:} First, we look at the current diagnostic confidence scores: pneumonia (0.68), upper respiratory infection (0.16), and pediatric diarrhea (0.16). Since pneumonia has the highest confidence, we prioritize its related symptoms. Among the candidate symptoms, nasal obstruction (0.267) is more common in pneumonia than listlessness (0.083) and sneezing (0.033). According to strategy 1, we should select the most typical symptom, so we choose "nasal obstruction" as the next symptom to ask the patient about. Final answer: choose nasal obstruction as the next symptom to inquire. \\
    \textbf{Symptom Status:} False \\
    \textbf{Response Reasoning:} To determine whether the patient is likely to have the symptom "nasal obstruction," we examine its frequency in pneumonia cases. The data shows a frequency of 0.267, which ranks 7th among all listed symptoms: lower than common ones like cough (0.883), sputum (0.65), and fever (0.533). Since "nasal obstruction" is not among the most typical symptoms of pneumonia, it is less likely the patient has it. Final judgment: False. \\
    \textbf{Diagnostic Confidence of Top Diseases:} \{"pneumonia": 0.66, "URTI": 0.18, "diarrhea": 0.16\} \\
    \hline
    
    \rowcolor{gray!15}
    \multicolumn{1}{l}{\textbf{Turn 7}} \\
    \textbf{Candidate Actions:} ["check for anorexia", "check for lethargy", "check for sneeze"] \\
    \textbf{Selected Symptom for Inquiry:} anorexia \\
    \textbf{Selection Reasoning:} The current diagnostic confidence for diseases is: Pneumonia: 0.66, Upper Respiratory Tract Infection: 0.18, Pediatric Diarrhea: 0.16. Since pneumonia has the highest confidence, we will focus on pneumonia-related symptoms. Among the candidate symptoms, anorexia (0.15), lethargy (0.083), and sneezing (0.117) are considered, with anorexia having the highest frequency. Based on Strategy 1, we will select anorexia as the next symptom to ask the patient. \\
    \textbf{Symptom Status:} True \\
    \textbf{Response Reasoning:} Anorexia is recorded in MCR, which indicates that the patient has this symptom. \\
    \textbf{Diagnostic Confidence of Top Diseases:} \{"pneumonia": 0.45, "URTI": 0.45, "diarrhea": 0.11\} \\
    \hline
    
    \rowcolor{gray!15}
    \multicolumn{1}{l}{\textbf{Turn 8}} \\
    \textbf{Candidate Actions:} ["check for sneeze", "check for pharyngeal discomfort", "check for lethargy"] \\
    \textbf{Selected Symptom for Inquiry:} sneeze \\
    \textbf{Selection Reasoning:} The diagnostic confidence for pneumonia and upper respiratory tract infection is the same (0.45), while pediatric diarrhea is much lower (0.11), so we can't apply Strategy 1. Instead, we use Strategy 2 and focus on symptoms related to the patient's condition. The patient's symptoms include cough, irritability, runny nose, fever, difficulty breathing, vomiting, phlegm, and anorexia. Looking at the frequency of candidate symptoms: Sneezing is more common in upper respiratory tract infection (0.233), while throat discomfort and lethargy appear less frequently across all diseases. Since sneezing is more related to upper respiratory tract infection and the patient's symptoms are mostly respiratory, sneezing helps distinguish between pneumonia and upper respiratory tract infection. Therefore, we should ask about sneezing next. \\
    \textbf{Symptom Status:} False \\
    \textbf{Response Reasoning:} To determine if the patient is likely to have the symptom `sneezing', we first look at its frequency in pneumonia and compare it to other symptoms. The frequency of sneezing in pneumonia is 0.117. When we rank all listed symptoms by frequency, sneezing is 13th with a relatively low frequency. The symptoms with higher frequencies include cough (0.883), phlegm (0.65), fever (0.533), and others. Given that sneezing ranks low in the list, it is not considered a typical symptom of pneumonia. Therefore, the conclusion is that the patient is unlikely to have the symptom `sneezing'. \\
    \textbf{Diagnostic Confidence of Top Diseases:} \{"pneumonia": 0.66, "URTI": 0.18, "diarrhea": 0.16\} \\
    \hline
    
    \rowcolor{gray!15}
    \multicolumn{1}{l}{\textbf{Turn 9}} \\
    \textbf{Candidate Actions:} ["check for pharyngeal discomfort", "check for lethargy"] \\
    \textbf{Selected Symptom for Inquiry:} lethargy \\
    \textbf{Selection Reasoning:} Given the diagnostic confidence for pneumonia (0.66), upper respiratory tract infection (0.18), and pediatric diarrhea (0.16), we prioritize symptoms related to pneumonia since its confidence is much higher. The candidate symptoms for pneumonia are lethargy (0.083) and throat discomfort (0.05). Although both have low frequencies, lethargy is more typical of pneumonia, so we choose it as the next symptom to ask the patient about. \\
    \textbf{Symptom Status:} False \\
    \textbf{Response Reasoning:} Lethargy is not recorded in MCR. This symptom is rarely present in pneumonia, so it is unlikely that the patient has this symptom. \\
    \textbf{Diagnostic Confidence of Top Diseases:} \{"pneumonia": 0.68, "URTI": 0.16, "HFMD": 0.16\} \\
    \hline
    
    \rowcolor{gray!15}
    \multicolumn{1}{l}{\textbf{Turn 10}} \\
    \textbf{Candidate Actions:} ["check for bubble-like spitting", "check for pulmonary shadow", "check for rash", "check for pharyngeal discomfort"] \\
    \textbf{Selected Symptom for Inquiry:} bubble-like spitting \\
    \textbf{Selection Reasoning:} Given the diagnostic confidence for pneumonia (0.68), upper respiratory tract infection (0.16), and hand-foot-and-mouth disease (0.16), we prioritize pneumonia-related symptoms since its confidence is significantly higher. The candidate symptoms for pneumonia are bubble-like spitting (0.167), lung shadow (0.167), throat discomfort (0.15), and rash (0.033). Both "bubble-like spitting" and "lung shadow" have the highest occurrence rates, but considering that bubble-like spitting may directly reflect lung dysfunction, it is likely more valuable. Therefore, we should choose "bubble-like spitting" as the next symptom to ask the patient about to further confirm pneumonia. \\
    \textbf{Symptom Status:} False \\
    \textbf{Response Reasoning:} To determine if the patient is likely to have the symptom "bubble-like spitting," we first check its frequency in pneumonia symptoms. The frequency is 0.167. When we rank all symptoms by frequency, "bubble-like spitting" ranks 9th. The most common symptoms of pneumonia include cough, sputum, and fever, while "bubble-like spitting" is relatively uncommon. Therefore, it’s not a typical pneumonia symptom. Based on this, we conclude that the patient is unlikely to have the symptom "bubble-like spitting." The judgment is False. \\
    \textbf{Diagnostic Confidence of Top Diseases:} \{"pneumonia": 0.68, "URTI": 0.16, "diarrhea": 0.16\} \\
    \hline

\caption{
A medical consultation case by DDO. The symptom status of True and False represent the presence and absence, respectively. "Top Diseases" refers to the diseases with the highest diagnostic confidence, ranked in the top-$w$ positions, with their diagnostic confidence values being processed using Softmax.
}
\label{tab:case_study}
\end{longtable}
}

\begin{table}[ht]
\centering
{\small
\begin{tabular}{p{0.96\linewidth}}
\hline
You are an experienced medical expert. The following information is provided to you:

~\\

\#\# Patient's Symptom Presentation

Present symptoms: \{positive\_symptoms\}

Absent symptoms: \{negative\_symptoms\}

\#\# Symptom Knowledge for \{candidate\_disease\}

Symptom occurrence frequencies based on existing diagnosed cases of \{candidate\_disease\}: \{empirical\_knowledge\}

~\\

Based on the patient's symptom presentation and the symptom knowledge of \{candidate\_disease\}, determine whether this disease can be diagnosed. The presence of relevant symptoms increases diagnostic confidence, while the absence of relevant symptoms decreases it. The influence of a symptom's presence or absence on diagnostic confidence increases with its typicality for the disease. There are two possible outputs: True or False. Output True if you believe the disease can be diagnosed; output False if it cannot.

Please output only the judgment result, without any additional content. \\
\hline
\end{tabular}}
\caption{The prompt used to estimate diagnostic confidence by BTP in our DDO framework.}
\label{tab:demo_disease_diagnosis_prompt}
\end{table}

\begin{table}[ht]
\centering
{\small
\begin{tabular}{p{0.96\linewidth}}
\hline
You are an experienced medical expert. Your task is to help select the next symptom to inquire about from a given set of candidate symptoms, in order to further collect the patient's symptom information. You are provided with the following information:

~\\

\# Current known patient symptom status (already inquired symptoms):

Present symptoms: \{positive\_symptoms\}

Absent symptoms: \{negative\_symptoms\}

\# Disease diagnostic confidence (confidence values range from 0 to 1; the higher the value, the more likely the disease):

\# Clinical presentation knowledge of diseases (symptom occurrence frequency based on historical case statistics): \{top\_diseases\_empirical\_knowledge\}

\# Candidate symptoms: \{candidate\_symptoms\}

~\\

Based on the above information, choose one suitable symptom from the candidate symptoms to ask next. The symptom selection strategies are as follows:

Strategy 1 (preferred): If the top-ranked disease has significantly higher diagnostic confidence than the others, choose a symptom from the candidate list that is relatively typical for the top-ranked disease, to help confirm its likelihood.

Strategy 2: If no candidate symptom fits Strategy 1, choose a symptom that is relatively related to the patient’s current symptom presentation.

Output format:

If there exists a suitable symptom `xx' in the candidate symptoms, output: Select `xx' as the next symptom to inquire about.

If no suitable symptom exists in the candidate symptoms, output: New candidate symptoms are needed.

Please think step by step. \\
\hline
\end{tabular}}
\caption{The prompt used to select an inquiry action in our DDO framework.}
\label{tab:demo_symptom_selection_prompt}
\end{table}

\begin{table}[ht]
\centering
{\small
\begin{tabular}{p{0.96\linewidth}}
\hline
You are a patient simulator. The disease that the simulated patient truly has is \{disease\_label\}.

~\\

The symptom knowledge of disease \{disease\_label\} is as follows:

Based on statistics from previously diagnosed cases of \{disease\_label\}, the symptom occurrence frequencies are: \{empirical\_knowledge\}

~\\

You need to determine whether the patient is likely to have the symptom \{inquiried\_symptom\} based on the symptom knowledge of the disease. The judgment should be either True or False:

True indicates that the patient is likely to have the symptom.

False indicates that the patient is unlikely to have the symptom.

~\\

The criteria for judgment are as follows:

If the symptom \{inquiried\_symptom\} is relatively typical for the disease \{disease\_label\} (i.e., it ranks high in the symptom occurrence frequency), then it is considered likely that the patient has the symptom (judgment should be True).

If the symptom is not typical for the disease, then the patient is considered unlikely to have it (judgment should be False).

Please think step by step and decide whether the patient is likely to have the symptom \{inquiried\_symptom\}. \\
\hline
\end{tabular}}
\caption{The prompt used to simulating the patient's response in our DDO framework.}
\label{tab:demo_patient_simulation_prompt}
\end{table}

\begin{table}[ht]
\centering
{\small
\begin{tabular}{p{0.96\linewidth}}
\hline
You are an experienced medical expert conducting a consultation with a patient. 

~\\

After several rounds of symptom inquiries, the patient has confirmed the following symptoms: \{positive\_symptoms\}. The patient has denied the following symptoms: \{negative\_symptoms\}. The diseases you suspect are: \{candidate\_diseases\}.

~\\

You need to decide whether to continue asking about symptoms to gather diagnostic evidence or to provide a final diagnosis based on the symptoms reported and your diagnostic knowledge of the diseases. The decision should be one of the following two options: (1) Ask about symptoms (2) Diagnose disease. Please provide your decision directly, without any additional explanation.

Decision: \\
\hline
\end{tabular}}
\caption{The prompt for deciding interaction action in the DP baseline.}
\label{tab:dp_interaction_decision_prompt}
\end{table}

\begin{table}[ht]
\centering
{\small
\begin{tabular}{p{0.96\linewidth}}
\hline
You are an experienced medical expert conducting a consultation with a patient. 

~\\

The symptoms that have already been inquired about and their status are as follows:  The symptoms confirmed by the patient: \{positive\_symptoms\}. The symptoms denied by the patient: \{negative\_symptoms\}.  

~\\

To increase diagnostic confidence, you need to choose a symptom to inquire about, ensuring that it has not been previously inquired about. Please provide the name of the symptom directly, without any additional content.  

The symptom to inquire about: \\
\hline
\end{tabular}}
\caption{The prompt for symptom inquiry in the DP baseline.}
\label{tab:dp_symptom_inquiry_prompt}
\end{table}

\begin{table}[ht]
\centering
{\small
\begin{tabular}{p{0.96\linewidth}}
\hline
You will play the role of a patient diagnosed with \{disease\}. 

~\\

Your symptom presentation is as follows: The symptoms you have: \{positive\_symptoms\}. The symptoms you do not have: \{negative\_symptoms\}.

~\\

Based on your symptom presentation, please answer truthfully whether you have the symptom \{symptom\}. The answer should be either `Yes' or `No'. Please provide the answer directly without any additional content.

Answer: \\
\hline
\end{tabular}}
\caption{The prompt for simulating the patient's response in the DP baseline.}
\label{tab:dp_simulating_patient_prompt}
\end{table}

\begin{table}[ht]
\centering
{\small
\begin{tabular}{p{0.96\linewidth}}
\hline
You are an experienced medical expert, currently conducting a consultation with a patient.

~\\

After several rounds of symptom inquiries: The symptoms confirmed by the patient are: \{positive\_symptoms\}. The symptoms denied by the patient are: \{negative\_symptoms\}. The diseases you suspect include: \{candidate\_diseases\}.

~\\

Please select three diseases from the suspected list above as the diagnosis results, ordered from most to least likely. Provide the answer in the form of a Python string list, and do not include any additional content.

Top three most likely diseases: \\
\hline
\end{tabular}}
\caption{The prompt for disease diagnosis in the DP baseline.}
\label{tab:dp_disease_diagnosis_prompt}
\end{table}

\end{CJK}
\end{document}